\documentclass[conference]{IEEEtran}
\IEEEoverridecommandlockouts
\usepackage{cite}
\usepackage{amsmath,amssymb,amsfonts}
\usepackage{algorithmic}
\usepackage{graphicx}             
\usepackage{textcomp} 
\usepackage{xcolor}

\usepackage{epsfig}                                                                             

\usepackage{booktabs}

\usepackage{url}

\usepackage{hyperref}

\def\BibTeX{{\rm B\kern-.05em{\sc i\kern-.025em b}\kern-.08em
    T\kern-.1667em\lower.7ex\hbox{E}\kern-.125emX}}
\begin{document}   

\title{CARE: Confidence-Aware Regression Estimation of building density fine-tuning EO Foundation Models
}    

%\author{\IEEEauthorblockN{Anonymous Authors}
%}   

\author{Nikolaos Dionelis, Jente Bosmans, Nicolas Longépé
\thanks{Nikolaos Dionelis, Jente Bosmans and Nicolas Longépé are with the European Space Agency (ESA), $\Phi$-lab. E-mail: Nikolaos.Dionelis@esa.int.}
}

\maketitle

%\textit{Motivation.}             
%\textit{Necessity.}  
%\textit{Evaluation.} 
\begin{abstract}   
Performing accurate confidence quantification and assessment in pixel-wise regression tasks, which are downstream applications of AI Foundation Models for Earth Observation (EO), is important for deep neural networks to predict their failures, improve their performance and enhance their capabilities in real-world applications, for their practical deployment. For pixel-wise regression tasks, specifically utilizing remote sensing data from satellite imagery in EO Foundation Models, confidence quantification is a critical challenge. The focus of this research is on developing a Foundation Model using EO satellite data that computes and assigns a confidence metric alongside regression outputs to improve the reliability and interpretability of predictions generated by deep neural networks. To this end, we develop, train and evaluate the proposed Confidence-Aware Regression Estimation (CARE) Foundation Model. Our model CARE computes and assigns confidence to regression results as downstream tasks of a Foundation Model for EO data, and performs a confidence-aware self-corrective learning method for the low-confidence regions. We evaluate the model CARE, and experimental results on multi-spectral data from the Copernicus Sentinel-2 constellation to estimate the building density (i.e. monitoring urban growth), show that the proposed method can be successfully applied to important regression problems in EO. We also show that our model CARE outperforms other methods.
\end{abstract}

\begin{IEEEkeywords}          
Earth Observation (EO), Pixel-wise regression, EO Foundation Models, Remote sensing, Copernicus Sentinel-2
%Earth Observation (EO), Remote sensing, EO Foundation Models, Pixel-wise regression.
\end{IEEEkeywords}

% https://www.themoonlight.io/en/review/care-confidence-aware-regression-estimation-of-building-density-fine-tuning-eo-foundation-models

\section{Introduction}               
%\textbf{Motivation.}   
\noindent The significance of confidence quantification and assessment in deep learning, specifically in AI Foundation Models in Earth Observation (EO) that use satellite data, for regression applications is critical. 
The utility of satellite data seems inexhaustible, and thanks to developments in AI, applications emerge at an accelerated pace in EO Foundation Models using remote sensing data.  
The \textit{trove}      of EO data is vast and expanding fast. The ESA Sentinel-2 constellation generates approximately 1.6TB of compressed multi-spectral data daily. 
However, the lack of labels stands in the way of \textit{accurately} performing important downstream tasks, and confidence quantification is critical to achieve better performance.
%The EO Foundation Models are pretrained on variable unlabeled satellite datasets that comprise large amounts of images, and special representations are extracted using self-supervised learning techniques. Then, in the downstream, the model is fine-tuned on a specific number of labeled data, and in the inference, classification, or regression are performed in context to the downstream task. 

%\textbf{Necessity.}    
The focus of this work is on developing methodological improvements in EO Foundation Models that: i) are needed to further improve performance for practicable real-world solutions that use EO data, and ii) are prerequisites to achieve important goals, namely \textit{accurately} performing tasks that are `downstream' from remote sensing data.   
The downstream tasks are, for example, building density estimation, land cover classification, crop type mapping and useful insights for Earth actions for sustainability and climate.             
Learning from \textit{unlabeled} data, i.e. devising \textit{new} methods for self-supervised learning, and developing an EO Foundation Model (FM) with confidence quantification and assessment are important. 
Confidence is a proxy for the probability of correct results. It is a metric between $0$ and $1$ that is an indicator of how well do we trust the output results of the model. Confidence is an \textit{a priori} estimate of the performance of our model.
\iffalse
Such tasks that are “downstream” from EO data require the acquisition of labels.  
Also, because Earth is dynamic, labels \textit{change} over time.

and having annotations for a geographical region at a specific point in time 
is not enough.
\fi

%For FMs, approximately $20\%$ of the labels for the use cases are needed, that would \textit{otherwise} be required.         
%By starting from a general model, FMs are able to perform building density estimation, as well as land cover mapping, road regression segmentation and crop type classification.

%Hence, c
%Confidence quantification and assessment is significant in Foundation Models on Earth Observation satellite data, particularly in regression contexts, i.e. buildings estimation downstream tasks, which are less frequently addressed compared to classification tasks. 
%For deep neural networks that solve regression problems, confidence estimation is very different from and more challenging than for neural network models that solve classification problems.        
%In addition, many more methods for confidence estimation have been proposed for classification neural networks, rather than for regression neural networks.    
The problem of accurately assigning a confidence metric to the real-valued output of a pixel-wise regression neural network in EO Foundation Models to estimate building density from satellite data is challenging and, to the best knowledge of the authors, has not been well addressed in the literature.        
%??? Because the softmax output layer is \textit{not} available in deep neural networks that solve regression problems, unlike classification neural networks \cite{Confidence1}, the interpretation, definition and meaning of confidence become \textit{less} clear \cite{Confquantification1} [8].         

\textbf{Our main contributions.}  
To address these problems, we propose the Confidence-Aware Regression Estimation (CARE) Foundation Model for remote sensing and EO.  
The model CARE computes and assigns confidence metrics alongside regression outputs in order to improve the reliability and interpretability of predictions generated by deep neural networks. 
The CARE model is implemented based on a modified U-Net architecture suitable for processing multi-spectral satellite images, which allows it to handle the complex spatial features present in the data.  
%The U-Net's encoder-decoder structure is particularly adept at tasks requiring precise localization, important for pixel-wise predictions in EO tasks.
Assigning a confidence metric to every inference of the model is crucial, as for the correctly estimated low-confidence samples in building density estimation from satellite data, our model subsequently implements a self-corrective method and improves its performance on them. 
CARE uses two heads that compute and assign a regression density measure and a confidence metric to the output of regression neural networks. 
Using our proposed loss function that comprises a Mean Squared Error (MSE) distance loss and a confidence loss, our model successfully performs self-corrective confidence-aware regression for building density estimation which is a downstream task of EO Foundation Models, and outperforms other methods.  
%CARE proposes a geospatial confidence-aware regression estimate, performs a proposed self-corrective ground truth-based method and recurrent self-fine-tuning to improve performance.  

Our main contributions are the development, training and evaluation of the proposed CARE model.
Results on multi-spectral Sentinel-2 data, and more specifically for the problem of building density estimation, which is the regression task of predicting how close buildings are to each other, how \textit{densely} built is the area in the image and how cities expand (i.e. urban growth) using EO Foundation Models, have shown that our method is effective and outperforms other base models.

\section{Related Work}                   
\noindent \textbf{Foundation Models in EO.} To use the vast amounts of unlabeled satellite data and achieve label efficiency, i.e. the ability to learn from only few labeled data in fine-tuning, EO Foundation Models are first pretrained on large unlabeled satellite datasets by performing self-supervised learning to extract \textit{special} representations from unlabeled data \cite{Prithvi1, newsatmaepp}. 
Then, in the downstream applications, Foundation Models are fine-tuned by performing supervised learning to specialize in specific applications, i.e. downstream tasks, by learning from a \textit{limited} number of labeled data.
In inference, classification or regression is performed for the downstream task \cite{Fibaek1, Prithvi1}.

\iffalse   
Such tasks that are “downstream” from EO data require the acquisition of labels.  
Also, because Earth is dynamic, labels \textit{change} over time.

and having annotations for a geographical region at a specific point in time 
is not enough. 
\fi 

Self-supervised learning and operating within an EO FM framework \cite{YiWangTUMPaper2025, PrithviIBMNASA} are motivated by the lack of labeled data and the technical impossibility to provide labels for the massive volume of data collected by the European Space Agency (ESA) and other space agencies (e.g., NASA, JAXA). 
Enhancing EO Foundation Models with confidence quantification and assessment is critical to improve their performance. 
%The trove of EO data collected by ESA Sentinel-2 constellation generates approximately 1.6TB of compressed data daily.   
%Lack of labels stands in the way of \textit{accurately} performing important downstream tasks, and confidence quantification is critical to achieve better performance.

\textbf{Confidence-aware models.}   
\noindent There are many different models that perform estimation and confidence assignment \cite{Confquantification2, regressionregressionnew2, regressionuseuseuse}. However, most of the existing models are not for pixel-wise regression building density estimation. Existing models for classification are based on the softmax output probability \cite{Confquantification1}.    
For regression problems where the result is a \textit{real-valued} number, the output softmax layer is not used in the neural network \cite{regressionregressionnew1}.               
Therefore, for regression neural networks that do not use softmax, it is \textit{not} straightforward to apply softmax-probability-based classification confidence methods.

%For classification problems, confidence estimation is commonly based on the softmax output layer and the maximum class probability. 
Without model retraining, it is possible to correct, as well as \textit{calibrate} to achieve no overfitting and use the \textit{entire} range of possible output values, the output softmax probability of deep neural networks \cite{Confidence1}. % that solve classification problems.            
For example, by using temperature scaling: Calibration based on entropy and the difference between the first and \textit{second} top class probabilities, or correction based on the aggregation of dispersion measures of softmax output probabilities over segments. 
Moreover, \textit{with} model retraining, two-headed networks can be used to estimate the \textit{True} Class Probability (TCP), which is a better confidence estimator than Maximum Class Probability (MCP) \cite{Confquantification4}.

For regression tasks, estimating model failures and regions of operation where the model should \textit{not} be trusted is challenging \cite{RegrConfi}.       
%It is important to address the shortcomings of the existing methods.              
%We also            
%Here, w 
We note that if we choose to turn a regression task into a classification problem \cite{regressionregressionnew2}, then we \textit{lose} precision, exactness and the desired outcome of having as an output a real-valued number. %, and having a result with an infinite number of decimal places.      

% In the literature, in addition to NLL and Bayesian neural networks, for confidence assignment and assessment in regression neural networks, evidential regression can also be performed, as well as conformal prediction (i.e. conformal inference) for regression. 

% 1/
% evidential

% 2/
% conformal

Considering the \textit{importance} of solving regression problems \cite{regressionregressionnew1}, for example in EO for the accurate estimation of canopy height Above Ground Biomass (AGB) \cite{Prithvi1}, or for the estimation of CO2 flux, it is crucial to develop methods that address the shortcomings of existing methods and \textit{correctly} compute and assign a confidence metric to the prediction outputs of regression neural networks \cite{Confquantification2}.

The work in \cite{newregression1} proposes a spatial self-corrective learning framework which: a) is able to \textit{locate}     predictions with potential large errors by estimating confidence using Maximum Likelihood Estimation (MLE), b) uses confidence-based pseudo-interpolation to correct low-confidence predictions in local neighborhoods, and c) performs recurrent self-refinement of the estimated height in an iterative manner.
%, and d) performs truth-based correction with a regression layer to address the challenges   
Experiments on different landscapes in the high-latitude regions show improvements compared to the other baseline methods.

The work in \cite{newregression2} 
% https://papers.ssrn.com/sol3/papers.cfm?abstract_id=4665951   
presents a probabilistic regression method. 
The deep neural network estimates the unnormalized density from the input-target pair using a conditional target density $p(y|x)$ model, with energy as the basis $(x, y)$. 
Monte Carlo sampling is used to reduce the negative log probability of this probabilistic $p(y|x)$ model. 
%Our technique outperforms linear regression and other probability- and confidence-based algorithms in four computer image regression challenges.  
%When Faster-RCNN is used for object recognition in the COCO dataset, the suggested model improves average accuracy (AP) by 2.3\% and provides a new, cutting-edge visual tracking system technique for bounding box prediction.  
%This technique covers more concerns than trust-based solutions, including age and head position.  
%This probabilistic regression method for computer vision is realistic and efficient.

Furthermore, in the literature \cite{newregression2, regressiontousenewuse, regressiontousenovel1, regressiontousenovel2}, in addition to probabilistic models and Bayesian neural networks, for confidence assignment and assessment in regression neural networks, evidential regression can also be performed \cite{regreesssiioonn1, regreesssiioonn2, regreesssiioonn3}, as well as conformal prediction (i.e. conformal inference) for regression \cite{reeggrreesssiioonn1, reeggrreesssiioonn2}.

\textbf{EO data noisy labels.}        
Most existing evaluation methods for EO Foundation Models \cite{Prithvi1, paperTerraMindIBM} are based \textit{only} on accuracy.              
However, remote sensing datasets might have noisy labels,  
i.e. incorrect annotations,   
and the evaluation of the performance of EO FMs using only the accuracy  
%on some specific EO datasets    
is not enough and might lead to misleading conclusions.   
Assessing the confidence with which the model performs inference is essential for \textit{usability} and uptake \cite{regressiioonn, SKondylatosPaper2025}.       
Accuracy, on its own, for the models and problems we consider, is \textit{not} a sufficient performance measure.            
An evaluation framework for FMs has to include confidence assessment to benchmark and compare different Foundation Models, evaluating their performance on both \textit{regression} and classification segmentation downstream tasks.

\iffalse     
In addition, for benchmarking and comparing EO FMs, a single combined evaluation metric has to include the performance accuracy on different \textit{diverse} downstream tasks, also including in the evaluation metric how diverse  
%these   
these use cases are, the number of labeled data in the $n$-shot downstream tasks, confidence quantification and assessment, the number of model weights retrained 
%(i.e. fine-tuning or freezing the FM weights)   
and the training epochs needed for convergence. 
%For the \textit{combined} evaluation metric, the individual metrics have to first be converted to a compatible format, standardized so that they have the same range of numerical values, and then integrated into a single metric.
\fi

\textbf{Confidence metric.}   
%\iffalse
The confidence metric helps humans to make correct decisions for critical matters \cite{Confquantification4, UQpaperforreggression, paperexplainabiityai1}, e.g. for policy.      
Moreover, to effectively combine possibly contradictory results from different models \cite{Confquantification1}, the assigned confidence metrics can be used to \textit{better}       inform the integration and fusion of the different model outputs.      
Within a Mixture of Experts (MoE) framework, we can effectively \textit{combine} different models using the confidence metric.     
Several different models might have outputs that lead to \textit{different} findings and conclusions.         
In this way, we are able to associate a confidence measure to the inference of each of the models in the MoE and, then, combine their outputs to the final model decision using the individual confidences.

Using the confidence metric that we have assigned to the model output, we are also able to perform anomaly detection, as low confidence is an \textit{indicator} of anomalies.       
This is important as anomaly detection/ Out-of-Distribution (OoD) detection is challenging, especially when the abnormalities are \textit{close} to the normal/ in-distribution samples in the image data space \cite{ADandOoD1, ADandOoD2}.               
In addition, the confidence metric also enables us to perform confidence-based change detection, which is crucial in scenarios when we do \textit{not} have labels for the change, as well as when the change is rare, occurring only a \textit{small} percentage of the time (e.g., $<5\%$, or even $<2\%$).

% Furthermore, in the literature \cite{newregression2, regressiontousenewuse, regressiontousenovel1, regressiontousenovel2}, in addition to probabilistic models and Bayesian neural networks, for confidence assignment and assessment in regression neural networks, evidential regression can also be performed \cite{regreesssiioonn1, regreesssiioonn2, regreesssiioonn3}, as well as conformal prediction (i.e. conformal inference) for regression \cite{reeggrreesssiioonn1, reeggrreesssiioonn2}.     

\begin{figure*}[tb]                                           
%\hspace{110pt}       
\centering \begin{minipage}[b]{.1\linewidth}              
  \centering                                                                    
  \centerline{\epsfig{figure=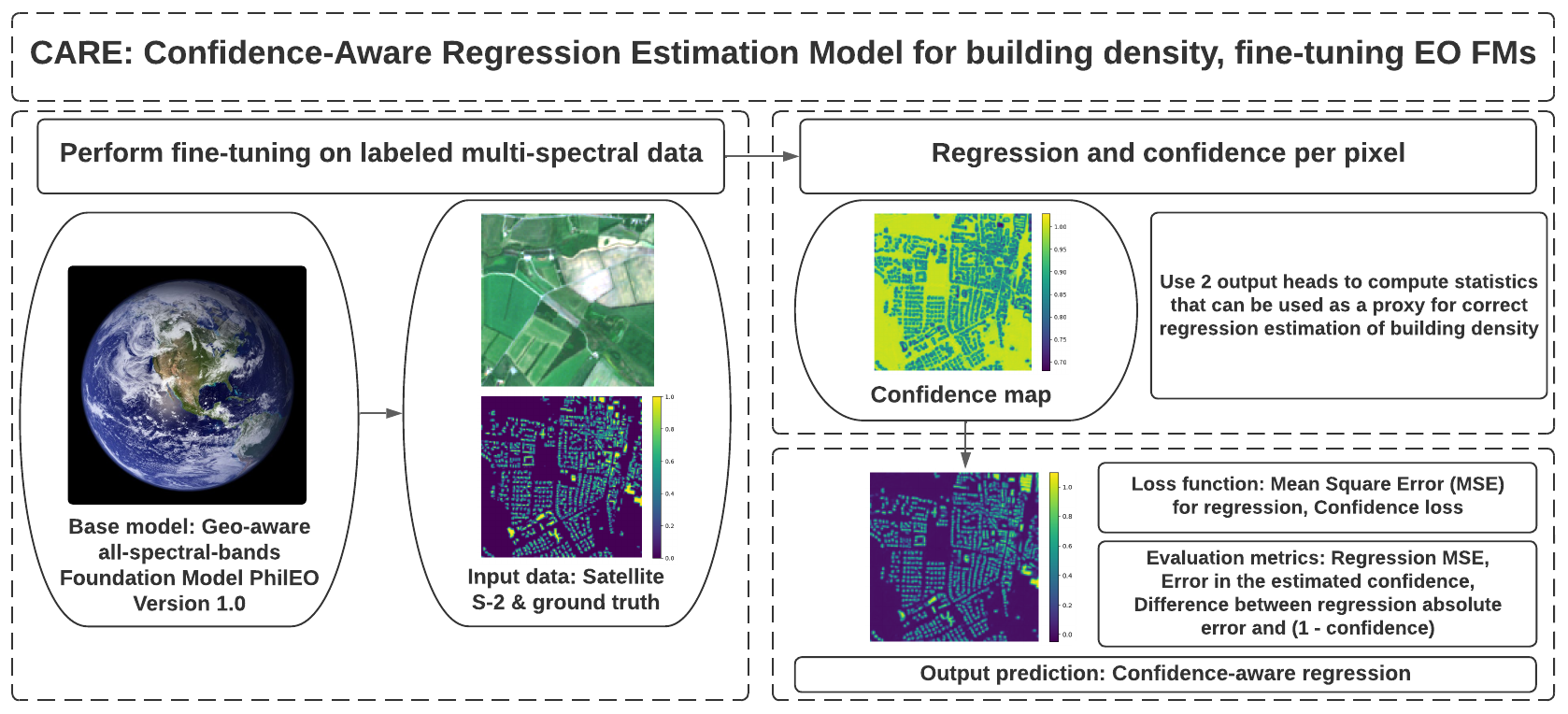,width=11.325cm}}                       
  \vspace{0.01cm}   
\end{minipage}      
%\hfill  
\vspace{-9pt}         
\caption{\small CARE for \textit{confidence}-aware regression: assign confidence metric to predictions, identify wrong predictions and refine {\color{black}the model.}}                  
%\caption{\small Flowchart diagram of the proposed model CAS that performs \textit{confidence}-aware segmentation where we assign a confidence metric to predictions, identify wrong predicted labels, and refine the model.}                   
\label{fig:flowchart}                                                      
\end{figure*}

%\fi   

\section{Our proposed methodology}\label{sec:secmethod}            
\noindent \textbf{Confidence-aware pixel-wise regression.}
\noindent To increase reliability and trust, accurately computing the confidence with which the model performs inference and outputs its results is desirable for neural networks that solve regression problems.          
For models to function and be operational in the real-world, estimating and assigning an \textit{accurate} confidence metric to every output model prediction is important.  
To successfully improve the performance of the model, assigning a confidence metric to every inference output of the model is crucial, as for the correctly estimated low-confidence samples, the model can subsequently \textit{improve}      its performance specifically on them.    
When we have samples with low confidence, we need to perform further retraining of the model on these samples or collect more data \textit{near} these samples (or data augmentation), in this region in the data space.          
Using the model CARE, we are able to achieve improved performance by assigning a confidence metric to the output results, and this is one of the main \textit{reasons} that confidence is important.          
Also, assigning a confidence metric to every output model prediction is essential, because for the \textit{low} confidence data, the model can choose to abstain from providing a prediction in these specific cases, rather than outputting an incorrect estimation result.

\subsection{Proposed method}       
\noindent The core methodology of the model CARE is based on a mini-batch training algorithm that takes advantage of the computed MSE distance measure assigned to the regression predictions.  
CARE uses two heads that compute and assign a regression building density measure and a confidence metric to the output of regression neural networks.
The approach is built upon sorting samples within mini-batches based on their prediction \textit{absolute} errors, which allows for establishing a relative confidence metric for each sample.

\textbf{Sorting mechanism.}    
In each mini-batch, samples are sorted in an ascending order of their errors (using the MSE as a measure of error). This enables the model to assign higher confidence scores to samples with lower errors and vice versa.

\textbf{Training procedure with dual outputs.} 
The model operates with \textit{two} output heads: one for the regression prediction and another for the confidence metric.   
The training of the model aims to minimize a custom loss function that consists of two components:  
a) Regression Loss (i.e. $L_0$): Measures the distance between the predicted outputs and the ground truth values using the MSE.    
b) Confidence Loss ($L_1$): Assesses the relationship between the prediction absolute errors and the assigned confidence scores, ensuring that higher confidence correlates with \textit{lower} prediction errors.

\textbf{Loss function.}        
The function minimized during training with Stochastic Gradient Descent (SGD) is given by:\begin{align}       
& \text{argmin}_{\pmb{\theta}_{nn}} \, L(\textbf{x}, \pmb{\theta}_{nn}, y, c, y^*, c^*)\text{,}\label{eq:newequ1}\\
& \text{where } L = L_0(\textbf{x}, \pmb{\theta}_{nn}, y, y^*) + \lambda \,  L_1(\textbf{x}, \pmb{\theta}_{nn}, y, c, y^*, c^*)\text{,}\label{eq:hasjdfjasx}
\end{align}
where we denote the neural network parameters of our model by $\pmb{\theta}_{nn}$. 
In the combined loss function \eqref{eq:newequ1}, $\lambda$ is a hyperparameter that can be adjusted to weigh the importance of confidence estimation relative to regression accuracy during training.
Here, the input image is $\textbf{x}_i$, the regression output $y_i$, the confidence output $c_i$, the ground truth regression value $y_i^*$ and the ground truth confidence $c_i^*$.     
The \textit{first} loss term, $L_0(\textbf{x}, \pmb{\theta}_{nn}, y, y^*)$, is the distance measure that is given by:\begin{align}   
& L_0(\textbf{x}, \pmb{\theta}_{nn}, y, y^*) = \dfrac{1}{N} \sum_{i=1}^{N} ||y_i - y_i^*||_2^2\text{,}\label{eq:equa1}
\end{align}
where we compute the MSE and use the batch size $N$.      
The second loss term, $L_1(\textbf{x}, \pmb{\theta}_{nn}, y, c, y^*, c^*)$, is for the estimated confidence, where $\forall i, j$: $c_i \geq c_j \Leftrightarrow d(y_i, y_i^*) \leq d(y_j, y_j^*)$. 
This second loss term is the \textit{confidence} metric that is given by:\begin{align}       
& L_1(\textbf{x}, \pmb{\theta}_{nn}, y, y^*, c, c^*) = \dfrac{1}{N} \sum_{i=1}^{N} |y_i - y_i^*|  \cdot      ||c_i - c_i^*||_2^2\text{.}\label{eq:confidequation} 
\end{align}

\textbf{Architecture.}          
The model CARE is implemented based on a modified U-Net architecture that has approximately $70$M parameters, suitable for processing multi-spectral satellite images in EO Foundation Models, which allows it to handle the complex spatial features present in the data \cite{Fibaek1}.   
The U-Net's encoder-decoder structure is particularly adept at tasks requiring precise localization, important for pixel-wise predictions in EO tasks.

\textbf{Datasets.}      
Our model CARE is pretrained on the PhilEO Globe Sentinel-2 global dataset \cite{Fibaek1, EGU1} and \textit{fine-tuned}     on the PhilEO Bench downstream tasks dataset \cite{Fibaek1}.  
CARE and the other examined baseline models in this work are evaluated on the PhilEO Bench \cite{Fibaek1, EGU1, Confidence1, Evaluating1}.

\subsection{CARE flowchart diagram}        
\noindent The proposed model CARE estimates the regression value and the confidence metric in the output of the deep neural network that solves pixel-wise regression problems in Fig.~\ref{fig:flowchart}.        
CARE assigns the confidence of $1$ for low error, and $0$ for \textit{high} error.     
%Using the model CARE, we are able to estimate confidence as an unbiased predictor of the \textit{error rate}.

%Here, for the implementation of the proposed methodology, we use the EO FM that we have recently trained, PhilEO \cite{Fibaek1, EGU1}.                
%Within the CARE framework, we are able to tackle and solve \textit{jointly} a group of problems (downstream tasks) that are of interest to the EO community \cite{Evaluating1}, \cite{Prithvi1}.           
For the training of the EO Foundation Model with unlabeled data, we have performed self-supervised learning that learns the correlations between: the Sentinel-2 multi-spectral images, masking and \textit{reconstruction}, geolocation (longitude and latitude) estimation and climate zone classes.      
%The architecture is a modified U-Net model \cite{Fibaek1}.          
For the downstream task of building density estimation, we fine-tune the EO Foundation Model, regress how close the buildings are in the image and evaluate the performance of the final model.

%The variation in the regression is modelled and captured by the estimated confidence metric of CARE.            
In this paper, we focus on the task of building density regression.     
%We have trained EO FMs using self-supervised learning, followed by supervised learning, making \textit{efficient} use of labels \cite{Evaluating1}, \cite{Fibaek1}.         
%Here, a 
%A methodology has been devised to take into account the geo-location longitude and latitude satellite information.        
%As a downstream task, we retrain the model to regress building density.       
To accurately perform confidence estimation, the network has \textit{two} output heads, i.e. one for the predicted regression output value and one for the estimated confidence, in \eqref{eq:equa1} and \eqref{eq:confidequation}.     
The latter is a confidence metric that correlates with the ground truth, indicating the extent to which we can \textit{trust} the output results of the other head, i.e. the predicted regression values \cite{Confquantification1, Confquantification2}.     

For $c_i^*$, in every mini-batch, after \textit{sorting} the samples in an ascending order with respect to their errors (i.e. MSE), $\eta = 80\%$ of the data are assigned with high confidence scores (i.e. $c_i^*=1$), while the \textit{remaining} 
data to low confidence scores (i.e. $c_i^*=0$). % \cite{Confquantification1}.        
%This is a ranking problem where correct predictions are ranked before incorrect ones.    
For the hyperparameter $\eta$, the threshold $80\%$ is chosen because we perform 
%a self-corrective method by 
model retraining, and we first set $\lambda=0$ in \eqref{eq:hasjdfjasx} and, then, set $\lambda > 0$ \cite{Confquantification4}, \cite{Confquantification1}. %and the proposed method is a two-round approach.              
%We note that we also sort the MSE or the absolute error.                   
%The results of CARE in Sec.~\ref{sec:sectionresultsuse} show that they are sensitive to small changes in the hyperparameter $\eta$, i.e. 20\% and 10\%.
%because of the first term in \eqref{eq:confidequation},      are \textit{not}      sensitive to relatively small changes in $\eta$. 
% a good model 
% two-round approach for fine-tuning  
During inference, for our \textit{decision rule}, to predict model failure, when the metric $(1-\text{confidence})$ is more than $\zeta = 20\%$ (or $10\%$) of the regression value result, then we have detected incorrect pixels and CARE performs self-correction by choosing to abstain from providing an output rather than predicting incorrect regression values.

\begin{table}[!tb]                                                                
    \caption{Evaluation of the proposed model CARE ($n=5000$), as well as comparison and ablation study, on the Sentinel-2 dataset PhilEO Bench Building Density Estimation (global, $10$ bands).}\label{tab:table2resultsCASmain2111}     
      %\vspace{5pt}        
      \centering                        
        \begin{tabular} 
{p{1.575cm} p{0.85cm} p{0.85cm} p{0.95cm} p{0.95cm} p{0.95cm}}    
  
\toprule            
    \normalsize {\small \textbf{Model}} &   
    \normalsize {\small \textbf{Error, Mean}} &  
    \normalsize {\small \textbf{Error, Med.}} & \normalsize {\small \textbf{MSE}} & \normalsize {\small \textbf{MSE, $20\%$}} & \normalsize {\small \textbf{MSE, $10\%$}}      
\\
\midrule                    
\midrule 
\normalsize {\small CARE (Ours)}        
& \normalsize {\normalsize  \small $\textbf{0.00759}$} &     
    \normalsize {\normalsize  \small $\text{0.00184}$}     & \normalsize {\normalsize \small $\textbf{0.00326}$}    & \normalsize {\normalsize \small $0.00057$}  & \normalsize {\normalsize \small $0.00033$}    
\\     
\midrule            
\midrule 
\normalsize {\small Gaussian-Output NLL} &   
\normalsize {\normalsize  \small $0.00997$} &   
\normalsize {\normalsize  \small $\textbf{0.00123}$}        & \normalsize {\normalsize  \small $0.00468$}    & \normalsize {\normalsize  \small $0.00053$}    & \normalsize {\normalsize  \small $0.00036$}      
\\   
\midrule           
\normalsize {\small Error Sorting \cite{Confquantification1}} &       
\normalsize {\normalsize  \small $0.10586$} &  
\normalsize {\normalsize  \small $0.03407$} & \normalsize {\normalsize  \small $\text{0.00334}$}  & \normalsize {\normalsize  \small $\textbf{0.00025}$}   & \normalsize {\normalsize  \small $\textbf{0.00025}$}      
\\ 
\midrule  
\normalsize {\small Absolute Error model} &   
\normalsize {\normalsize  \small $0.00847$} & 
\normalsize {\normalsize  \small $0.00254$}   & \normalsize {\normalsize  \small $0.00329$} & \normalsize {\normalsize  \small $0.00056$}  & \normalsize {\normalsize  \small $0.00032$}     
\\     
\midrule                                         
\midrule  
\end{tabular}               
\end{table}

\iffalse
Using the confidence metric we have assigned to the model output, we are also able to perform anomaly detection, as low confidence is an \textit{indicator} of anomalies.     
This is important as anomaly detection/ Out-of-Distribution (OoD) detection is challenging, especially when the abnormalities are \textit{close} to the normal/ in-distribution samples in the image data space \cite{ADandOoD1, ADandOoD2}.            
Confidence also enables us to perform confidence-based change detection, which is crucial in scenarios when we do \textit{not} have labels for the change, as well as when the change is rare, occurring only a \textit{small} percentage of the time ($\leq5\%$).

The confidence metric helps humans to make correct decisions for critical matters, e.g. for policy.      
Moreover, to effectively combine possibly contradictory results from different models, the assigned confidence metrics can be used to better inform the fusion and integration of the different model outputs.    
Within a Mixture of Experts (MoE) framework, we can effectively \textit{combine} different models using the confidence metric.  
The several different models might have outputs that lead to \textit{different} findings and conclusions.      
In this way, we are able to associate a confidence measure to the inference of each of the models in the MoE and, then, combine their outputs to the final model decision using the individual confidences.
\fi

\section{Evaluation and Results}\label{sec:sectionresultsuse}                                     
\noindent The evaluation of the proposed model CARE is based on the observation that the absolute error of the regression output should be equal to the quantity: $(1-\text{confidence})$.             
This should be true for every pixel and image.         
%In this way, we numerically evaluate the performance our model.       
We examine if we learn to \textit{both}: predict building density, and estimate confidence.     
We compute the average absolute error between: i) the absolute error of the regression output, and ii) the metric $(1-\text{confidence})$.            
The mean is over the pixels and the images, and the \textit{ideal} result is zero.        
In the evaluation results, we are interested in the error between the regression absolute error and the metric $(1-\text{confidence})$ being \textit{very small}.       
Therefore, in this work, for the evaluation of the proposed model CARE and the estimated confidence, we use the MSE.
%Mean Absolute Error (MAE), and      we also calculate the square Root MSE (RMSE).         
Furthermore, we compute the median absolute error, i.e. using the \textit{median} instead of the average, where the median is more robust to outliers than the mean.         
The median absolute error, in our case for building density estimation, is much lower than the average error.

\subsection{Evaluation of CARE using absolute error in Table I and comparison to other models}  
\noindent In Table~\ref{tab:table2resultsCASmain2111}, we evaluate the proposed model CARE and examine both the average error and the \textit{median} error for the results of the model.   
The median error of the proposed model CARE is low, 
i.e. $0.00184$, and        
this is desirable as it shows that the difference between the absolute error of the \textit{regression}  output and the estimated metric $(1-\text{confidence})$ is small.                  
We also evaluate the regression output result using the MSE, %like in \eqref{eq:equa1}, 
in Tables~\ref{tab:table2resultsCASmain2111} and \ref{tab:table2resultsCASmainn22222}.              
We run experiments using $n$-sample tests, %similar to \cite{Fibaek1}, 
where we use $n$ training samples per region for fine-tuning.                
Stratified sampling is performed, %as in the PhilEO Bench, 
and the different \textit{regions} are: Denmark, East Africa, Egypt, Guinea, Europe, Ghana, Israel, Japan, Nigeria, North America, Senegal, South America, Tanzania and Uganda \cite{Fibaek1, Evaluating1}.              
The performance of the model CARE, for $n=5000$, is examined in Table~\ref{tab:table2resultsCASmain2111}.

\begin{table}[!tb]                                                     
    \caption{Evaluation of the proposed model CARE on the dataset PhilEO Bench Building Density Estimation (Sentinel-2 data, L2A).}\label{tab:table2resultsCASmainn22222}    
      %\vspace{5pt}                   
      \centering                            
        \begin{tabular} 
{p{0.56cm} p{0.73cm} p{0.73cm} p{0.73cm} p{0.64cm} p{0.64cm} p{0.64cm} p{0.79cm}}          
  
\toprule            
    \normalsize {\small ~~~~ $n$=}     &        
    \normalsize {\small \textbf{10000}} &      
    \normalsize {\small \textbf{7500}} &     
    \normalsize {\small \textbf{5000}} &  
    \normalsize {\small \textbf{1000}} & \normalsize {\small \textbf{500}} & \normalsize {\small \textbf{100}} & \normalsize {\small \textbf{50}}         
\\     
\midrule                   
\midrule 
\normalsize {\small \textbf{Error, Mean}}           
& \normalsize {\normalsize  \small $\text{0.00683}$}     & \normalsize {\normalsize  \small $\text{0.00761}$}    &    \normalsize {\normalsize  \small $\text{0.00759}$}  &           
    \normalsize {\normalsize  \small $\text{0.0138}$}     &  \normalsize {\normalsize \small $0.0167$}     & \normalsize {\normalsize \small $0.0246$}    & \normalsize {\normalsize \small $0.0261$}     
\\     
\midrule         
%\midrule 
\normalsize {\small \textbf{Error, Med.}}  &   
\normalsize {\normalsize  \small $0.00150$}     &    
\normalsize {\normalsize  \small $0.00190$}    &   
\normalsize {\normalsize  \small $0.00184$} & 
\normalsize {\normalsize  \small $0.0065$}     & \normalsize {\normalsize  \small $\text{0.0088}$}     & \normalsize {\normalsize  \small $\text{0.0147}$}    &  \normalsize {\normalsize  \small $\text{0.0159}$}    
\\   
\midrule    
\midrule       
\normalsize {\small \textbf{MSE}} &   
\normalsize {\normalsize  \small $0.00301$}     &   
\normalsize {\normalsize  \small $0.00316$} &    
\normalsize {\normalsize  \small $0.00326$} & 
\normalsize {\normalsize  \small $0.0039$}    & \normalsize {\normalsize  \small $\text{0.0043}$}     & \normalsize {\normalsize  \small $\text{0.0051}$}     & \normalsize {\normalsize  \small $\text{0.0034}$}      
\\  
\midrule                                   
\normalsize {\small \textbf{MSE, $20\%$}} &   
\normalsize {\normalsize  \small $0.00062$}    &    
\normalsize {\normalsize  \small $0.00054$}    &   
\normalsize {\normalsize  \small $0.00057$} &  
\normalsize {\normalsize  \small $0.0005$}  & \normalsize {\normalsize  \small $\text{0.0006}$}    & \normalsize {\normalsize  \small $\text{0.0011}$}    & \normalsize {\normalsize  \small $\text{0.0011}$}     
\\ 
\midrule
\normalsize {\small \textbf{MSE, $10\%$}} &    
\normalsize {\normalsize  \small $0.00031$}    &   
\normalsize {\normalsize  \small $0.00029$}    &  
\normalsize {\normalsize  \small $0.00033$} & 
\normalsize {\normalsize  \small $0.0004$}     & \normalsize {\normalsize  \small $\text{0.0005}$}    & \normalsize {\normalsize  \small $\text{0.0010}$}    & \normalsize {\normalsize  \small $\text{0.0007}$}      
\\      
\midrule         
\midrule  
\end{tabular}             
\end{table}

For the evaluation of CARE, we also compare our model with other models. 
In Table~\ref{tab:table2resultsCASmain2111}, we \textit{compare} the results we obtain when we use Negative Log Likelihood (NLL) \cite{RegrConfi} and the model Gaussian-Output NLL.    
The percentage improvement of CARE compared to Gaussian-Output NLL is 
$30.34\%$   
for the MSE, and  
$23.87\%$   
for the average error. %, where $(\text{final}-\text{initial}) / \text{initial}$.     
We observe that according to these results, for the use case of building density estimation, regarding the model Gaussian-Output NLL, an interesting open research question is how useful the normal distribution assumption is.   
Moreover, we also compare CARE with the model Error-Sorting Confidence \cite{Confquantification1} in Table~\ref{tab:table2resultsCASmain2111}.     
The percentage improvement of our model compared to Error-Sorting Confidence is  
$2.40\%$     
for the MSE, and  
$92.83\%$    
for the mean absolute error.     
%Also, t

The percentage improvement of CARE compared to the model Absolute-Error Confidence  
is   
$0.91\%$     
for the MSE, and 
$10.39\%$
for the average error.     
The Absolute-Error Confidence is also an ablation study, using only the \textit{first} term in \eqref{eq:confidequation}.     
These results demonstrate the efficacy of 
%the proposed 
our 
algorithm and the superiority of 
%our approach   
CARE 
in outputting a reliable confidence metric for pixel-wise regression tasks, leading to 
both trustworthy and accurate model predictions.

\begin{table}[!tb]                                                     
    \caption{Evaluation of the model Error-Sorting Confidence \cite{Confquantification1} on the PhilEO Bench building density regression, for comparison with the results of CARE in Table~\ref{tab:table2resultsCASmainn22222}, where Error Sorting is also an ablation study, i.e. using \textit{only}       the second term in \eqref{eq:confidequation}.}\label{tab:table2resultsCASmain22333}\label{tab:table2resultsCASmainn222223}    
      %\vspace{5pt}                    
      \centering                           
        \begin{tabular} 
{p{0.56cm} p{0.73cm} p{0.73cm} p{0.73cm} p{0.64cm} p{0.64cm} p{0.64cm} p{0.79cm}}          
  
\toprule             
    \normalsize {\small ~~~~ n=}    &         
    \normalsize {\small \textbf{10000}} &      
    \normalsize {\small \textbf{7500}} &     
    \normalsize {\small \textbf{5000}} &  
    \normalsize {\small \textbf{1000}} & \normalsize {\small \textbf{500}} & \normalsize {\small \textbf{100}} & \normalsize {\small \textbf{50}}         
\\     
\midrule                    
\midrule 
\normalsize {\small \textbf{Error, Mean}}           
& \normalsize {\normalsize  \small $\text{0.09969}$}   & \normalsize {\normalsize  \small $\text{0.08420}$}   & \normalsize {\normalsize  \small $\text{0.10586}$}      &         
    \normalsize {\normalsize  \small $\text{0.0861}$}       & \normalsize {\normalsize \small $0.0981$}        & \normalsize {\normalsize \small $0.0963$}     & \normalsize {\normalsize \small $0.1316$}       
\\   
\midrule            
%\midrule   
\normalsize {\small \textbf{Error, Med.}}  &   
\normalsize {\normalsize  \small $0.02332$}      &   
\normalsize {\normalsize  \small $0.02185$}   &  
\normalsize {\normalsize  \small $0.03407$}        & 
\normalsize {\normalsize  \small $0.0323$}       & \normalsize {\normalsize  \small $\text{0.0417}$}       & \normalsize {\normalsize  \small $\text{0.0554}$}          &  \normalsize {\normalsize  \small $\text{0.0765}$}        
\\    
\midrule    
\midrule       
\normalsize {\small \textbf{MSE}} &   
\normalsize {\normalsize  \small $0.00307$}       &  
\normalsize {\normalsize  \small $0.00320$}   & 
\normalsize {\normalsize  \small $0.00334$}        & 
\normalsize {\normalsize  \small $0.0039$}      & \normalsize {\normalsize  \small $\text{0.0042}$}        & \normalsize {\normalsize  \small $\text{0.0056}$}      & \normalsize {\normalsize  \small $\text{0.0035}$}         
\\      
\midrule                                   
\normalsize {\small \textbf{MSE, $20\%$}} &   
\normalsize {\normalsize  \small $0.00024$}     &   
\normalsize {\normalsize  \small $0.00027$}   &  
\normalsize {\normalsize  \small $0.00025$}  &  
\normalsize {\normalsize  \small $0.0004$}    & \normalsize {\normalsize  \small $\text{0.0005}$}      & \normalsize {\normalsize  \small $\text{0.0013}$}        & \normalsize {\normalsize  \small $\text{0.0004}$}        
\\     
\midrule     
\normalsize {\small \textbf{MSE, $10\%$}}  &     
\normalsize {\normalsize  \small $0.00024$}  &  
\normalsize {\normalsize  \small $0.00027$}    & 
\normalsize {\normalsize  \small $0.00025$}  & 
\normalsize {\normalsize  \small $0.0004$}           & \normalsize {\normalsize  \small $\text{0.0005}$}        & \normalsize {\normalsize  \small $\text{0.0013}$}      & \normalsize {\normalsize  \small $\text{0.0004}$}         
\\               
\midrule                       
\midrule  
\end{tabular}                  
\end{table}

We therefore have compared CARE to the Gaussian-Output NLL model in Table~\ref{tab:table2resultsCASmain2111}, where the \textit{latter}      minimizes the Gaussian negative logarithmic likelihood loss. %(i.e. in PyTorch, {gaussian\_nll\_loss}).       
Hence, the MSE loss for the regression error achieves \textit{higher} accuracy (i.e. $0.00326$ compared to $0.00468$), and this is one of our \textit{main} targets. 
For the evaluation of CARE, in addition to the average and median absolute errors, we also compute the correlation between the confidence
and the error in the regression. %output. %, similar to \cite{Confidence1}.       
%The correlation coefficient, being a ratio, is independent of the units of measurement of the two quantities, i.e. confidence metric and regression absolute error, and this is desirable.                
The correlation is $0.62090$. For the model Gaussian-Output NLL, it
is $0.56696$. Here, the percentage improvement is $9.51\%$.

We also compare the proposed model CARE to the ensembles method with $M=\{3, 1\}$ members \cite{EnsembleNew}.                     
By combining the \textit{Gaussian} distributions from the ensemble members using a heteroscedastic Gaussian log-likelihood loss, this method estimates the epistemic uncertainty.             
For $n=1000$, for $M=3$, the MSE and mean absolute error are $0.00685$ and $0.01306$, respectively, while for $M=1$, these are $0.00696$ and $0.01353$.   
For CARE, in Table~\ref{tab:table2resultsCASmainn22222}, these are $0.0039$ and $0.0138$.            
Here, because of \eqref{eq:hasjdfjasx} and \eqref{eq:equa1}, the percentage improvement in MSE of our model CARE, in comparison to the $M=3$ ensembles method \cite{EnsembleNew}, is $43.1\%$.

\subsection{Further evaluation of CARE based on threshold $\zeta$, Table I} 
\noindent Furthermore, for the evaluation of CARE, we also examine in Table~\ref{tab:table2resultsCASmain2111} the MSE $20\%$ or $10\%$ which is a threshold, i.e. $\zeta$, to detect instances where the model simply does \textit{not} know the correct result (i.e. pixel-wise regression value) from the available input data, e.g. due to lack of spectral information or resolution.     
In such cases, models might choose to abstain from providing an answer and should be able to output ``\textit{None} of the above'' for the result value of segmentation.      
A building density of $20\%$ with an estimated error of $\pm 10\%$ leads to the predicted density of $20\%$ being \textit{not} useful, i.e. not accurate.          
In contrast, a building density of $80\%$ with an estimated error of $\pm 1\%$ leads to the predicted density of $80\%$ being useful and accurate.   
%Here, a     

\begin{figure}[tb]                                                              
\hspace{26pt}     
\begin{minipage}[b]{.1\linewidth}          
  \centering      
  \centerline{\epsfig{figure=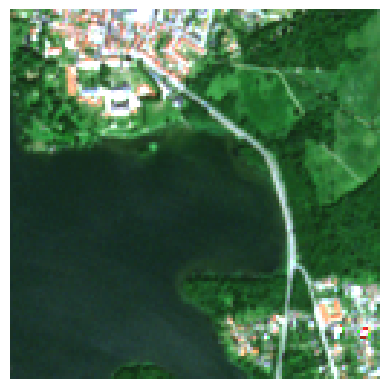,height=2.945cm,width=2.945cm}}           
  \vspace{0.001cm}   
  \centerline{\small a) Input image~~}\medskip  
\end{minipage}
%\hfill
\hspace{49pt} 
\begin{minipage}[b]{0.1\linewidth}
  \centering 
  \centerline{\epsfig{figure=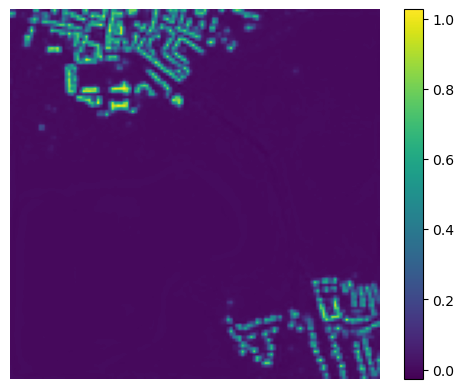,height=2.945cm,width=2.945cm}}      
  \vspace{0.001cm} 
  \centerline{\small b) Prediction~~}\medskip  
\end{minipage}
%\hfill 
\hspace{49pt}    
\begin{minipage}[b]{.1\linewidth}    
  \centering      
  \centerline{\epsfig{figure=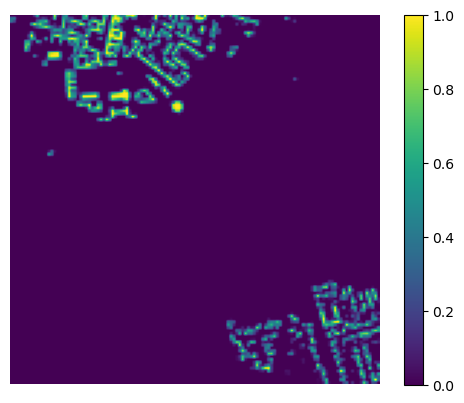,height=2.945cm,width=2.945cm}}      
  \vspace{0.001cm}    
  \centerline{\small c) Ground truth~~}\medskip 
\end{minipage}              

\hspace{26pt}   
\begin{minipage}[b]{.1\linewidth}          
  \centering         
  \centerline{\epsfig{figure=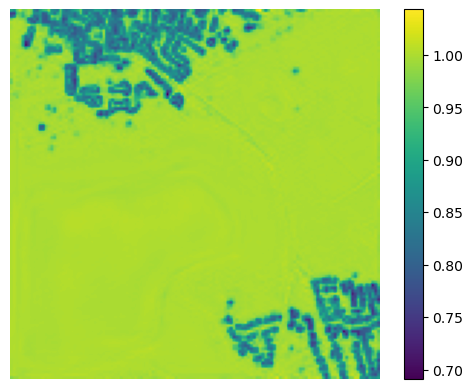,height=2.945cm,width=2.945cm}}    
  \vspace{0.001cm}  
  %\centerline{\small d) Incorrect~~}
  %\centerline{\small \, classifications~~}\medskip  
  \centerline{\small d) Confidence~~~}      
  \centerline{\small \, map, CARE (Ours)~~~}\medskip     
\end{minipage}
%\hfill
\hspace{49pt}
\begin{minipage}[b]{0.1\linewidth}   
  \centering         
  \centerline{\epsfig{figure=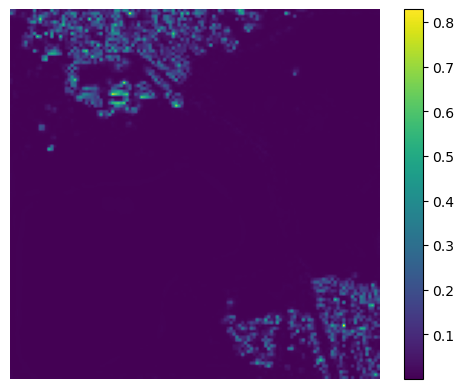,height=2.945cm,width=2.945cm}}       
  \vspace{0.001cm}   
  \centerline{\small e) Abs. error btw~~~}      
  \centerline{\small prediction \& gt~~~}\medskip   
\end{minipage} 
%\hfill  
\hspace{49pt}    
\begin{minipage}[b]{.1\linewidth}                
  \centering                        
  \centerline{\epsfig{figure=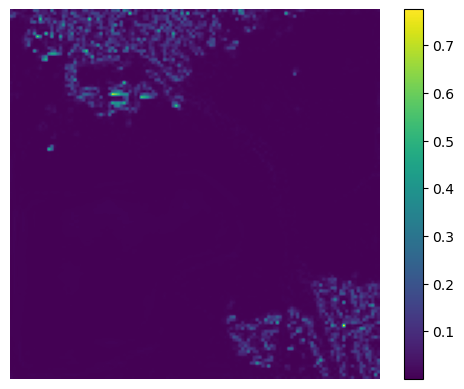,height=2.945cm,width=2.945cm}}                   
  \vspace{0.001cm}     
  \centerline{\small f) Abs. error btw~}           
  \centerline{\small pred. uncertainty \& (e)~}\medskip        
\end{minipage}    
\vspace{-9pt}         
%\caption{\small Semantic segmentation and confidence assignment and assessments by CAS on Sentinel-2 \textit{multi-spectral} data using WorldCover.}                                                              
%\caption{\small Semantic segmentation and confidence assignment and assessments by CAS on Sentinel-2 \textit{multi-spectral} data using WorldCover.}                                                
\caption{\small Pixel-wise regression and confidence estimation and assessment by the model CARE described in Sec.~\ref{sec:secmethod} on Sentinel-2 data.}                                                   
%Semantic segmentation, confidence estimation, and confidence assessments by our model CAS described in Sec.~\ref{sec:sectionProposedMethod} in the subsection ``Model CAS'' on Sentinel-2 images using the dataset \textit{WorldCover}      
\label{fig:figure4CAS}                               
\label{fig:figure3CAS}    
\end{figure}

A prediction of the model with a \textit{high} confidence indicates high reliability and trust for this particular prediction.                      
Also, the features for high building density might \textit{not} be clear in multi-spectral optical EO data.        
This is due to epistemic uncertainty, as it is induced by the lack of detail in the measurement.    
This is why a plausible output set that is an \textit{Open Set}, is needed.            
Epistemic uncertainty is systematic, caused by lack of knowledge, and can be reduced by learning the characteristics of the quantity (e.g., high building density) using additional information (for example, \textit{in-situ} measurements).       
Moreover, \textit{aleatoric}       
uncertainty is statistical and related to randomness, and the sample not being a typical example of the quantity (e.g., very high or \textit{low} building density).

\begin{figure}[tb]                                                         
\hspace{26pt}    
\begin{minipage}[b]{.1\linewidth}         
  \centering      
  \centerline{\epsfig{figure=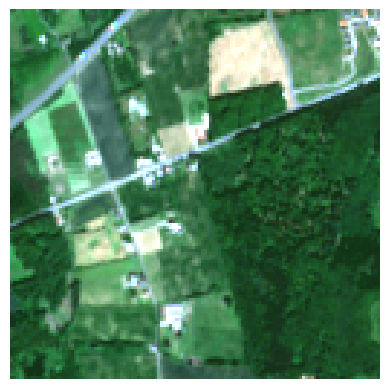,height=2.945cm,width=2.945cm}}           
  \vspace{0.001cm}   
  \centerline{\small a) Input image~~}\medskip  
\end{minipage}
%\hfill
\hspace{49pt}
\begin{minipage}[b]{0.1\linewidth}
  \centering 
  \centerline{\epsfig{figure=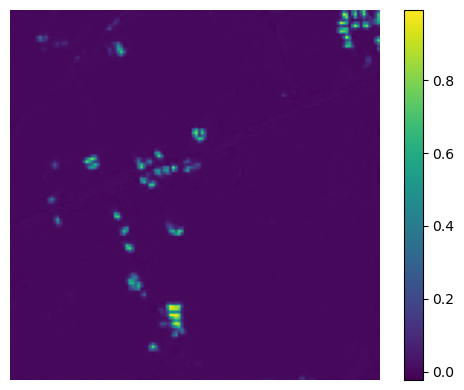,height=2.945cm,width=2.945cm}}      
  \vspace{0.001cm} 
  \centerline{\small b) Prediction~~}\medskip  
\end{minipage}
%\hfill 
\hspace{49pt}    
\begin{minipage}[b]{.1\linewidth}    
  \centering      
  \centerline{\epsfig{figure=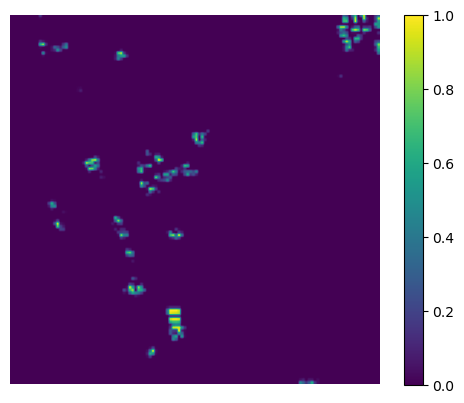,height=2.945cm,width=2.945cm}}      
  \vspace{0.001cm}    
  \centerline{\small c) Ground truth~~}\medskip 
\end{minipage}              

\hspace{26pt}   
\begin{minipage}[b]{.1\linewidth}          
  \centering         
  \centerline{\epsfig{figure=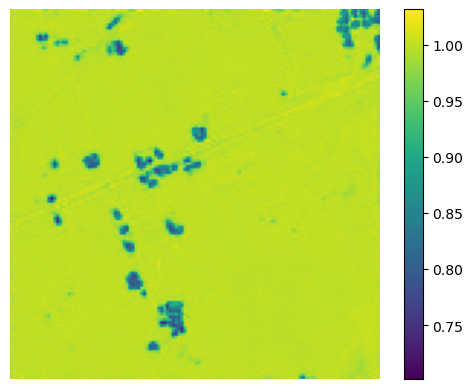,height=2.945cm,width=2.945cm}}    
  \vspace{0.001cm}  
  %\centerline{\small d) Incorrect~~}
  %\centerline{\small \, classifications~~}\medskip  
  \centerline{\small d) Confidence~~~}      
  \centerline{\small \, map, CARE (\textit{Ours})~~~}\medskip    
\end{minipage}
%\hfill
\hspace{49pt}
\begin{minipage}[b]{0.1\linewidth}   
  \centering         
  \centerline{\epsfig{figure=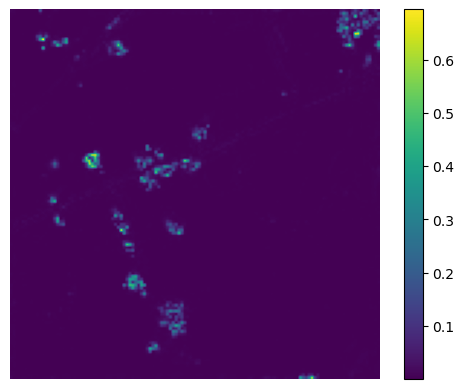,height=2.945cm,width=2.945cm}}       
  \vspace{0.001cm}   
  \centerline{\small e) Abs. error btw~~~}      
  \centerline{\small prediction \& gt~~~}\medskip   
\end{minipage} 
%\hfill  
\hspace{49pt}    
\begin{minipage}[b]{.1\linewidth}                
  \centering                        
  \centerline{\epsfig{figure=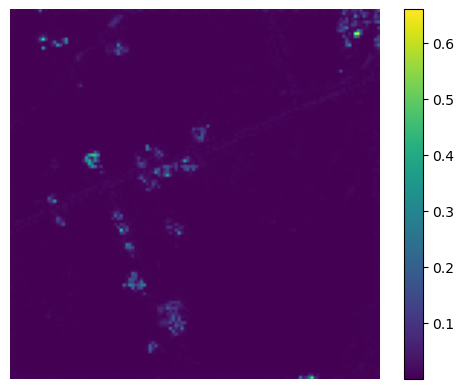,height=2.945cm,width=2.945cm}}                    
  \vspace{0.001cm}     
  \centerline{\small f) Abs. error btw~}           
  \centerline{\small pred. uncertainty \& (e)~}\medskip        
\end{minipage}    
\vspace{-9pt}        
%\caption{\small Semantic segmentation and confidence assignment and assessments by CAS on Sentinel-2 \textit{multi-spectral} data using WorldCover.}                                                              
%\caption{\small Semantic segmentation and confidence assignment and assessments by CAS on Sentinel-2 \textit{multi-spectral} data using WorldCover.}                                                    
\caption{\small CARE regression and confidence estimation: Building density.}                                                        
%Semantic segmentation, confidence estimation, and confidence assessments by our model CAS described in Sec.~\ref{sec:sectionProposedMethod} in the subsection ``Model CAS'' on Sentinel-2 images using the dataset \textit{WorldCover}     
\label{fig:figure4CAS2}                             
\label{fig:figure3CAS2}   
\end{figure}

Our experiments in Tables~\ref{tab:table2resultsCASmain2111}-\ref{tab:table2resultsCASmain22333} show that CARE is effective   
%, 
%its ablation study is a success,     
and  
%our model  
outperforms other 
%baseline     
models.           
The focal point of this paper is confidence quantification and assessment for pixel-wise regression tasks in EO using neural networks that have a continuous output. %(real-valued).      
%We have used EO FMs, and as a downstream task, we have performed building density estimation using the PhilEO Bench \cite{Fibaek1, EGU1, Evaluating1}.
%We have implemented CARE in PyTorch and we have developed, trained and evaluated our model.   
%For our experiments, we have used a A100 SXM4 40 GB GPU,  as well as 4 A100 SXM4 40 GB GPUs.        
%During training, the learning curve loss value decays quickly, and the errors are already imperceptible after $20$-$30$ epochs.          
CARE achieves good generalisation performance, and this work's methodological and model development value, as well as its application implementation value, is high.     
The obtained results can be \textit{useful} for researchers and practitioners (i.e. from a practical viewpoint).                          
Using a mathematical definition for the confidence metric for pixel-wise regression tasks, a contribution of this paper is the value for \textit{applications}, so that researchers studying such \textit{real-world} problems can take advantage of the results and the attained good performance.

\subsection{Evaluation of CARE using n-shot experiments in Table II}
\noindent We evaluate the proposed model CARE in Table~\ref{tab:table2resultsCASmainn22222} using $n$-sample tests, for $n$ from $50$ to $10000$ samples.        
We examine both the MSE for the accuracy of the regression, i.e. \eqref{eq:equa1} in \eqref{eq:hasjdfjasx}, and the average error for the correctness of the confidence metric, i.e. $c_i$ in \eqref{eq:confidequation}.    
Here, we observe that for $n=7500$ samples, the MSE is $0.00316$, while for $n=500$, the MSE is $0.0043$, i.e. the performance of the model \textit{improves} when we have more labeled data.    
In addition, for the accuracy of the confidence metric of our model, the median error in Table~\ref{tab:table2resultsCASmainn22222} is $0.00190$ for $n=7500$ samples, while the median error is $0.0088$ for $n=500$.      
The performance of the model CARE, for confidence quantification, improves when we have more labeled data, i.e. $n=7500$ samples compared to $n=500$.

For $n=1000$ in Table~\ref{tab:table2resultsCASmainn22222}, CARE achieves the MSE of $0.0039$, for the accuracy of the building density regression.       
The model Gaussian-Output NLL from Table~\ref{tab:table2resultsCASmain2111}, for $n=1000$, yields the MSE of $0.0052$.         
Here, the percentage improvement of our model, compared to Gaussian-Output NLL, is $25\%$.

%\subsection{Further evaluation of CARE in Table II}  
For the evaluation of our model CARE, we also examine in Table~\ref{tab:table2resultsCASmainn22222} the MSE $20\%$ or $10\%$, similar to in Table~\ref{tab:table2resultsCASmain2111}.  
We observe that the performance of the proposed model CARE for estimating the building density, i.e. how close the buildings are in the Sentinel-2 \textit{multi-spectral} images, improves when the model has the ability to abstain in cases of \textit{low} confidence. 
Moreover, Table~\ref{tab:table2resultsCASmainn22222} shows that when assuming that $10\%$ provide low confidence and our model should abstain, CARE gives improved performance, compared to the assumption of $20\%$.
These results demonstrate the effectiveness of our algorithm in providing a reliable confidence metric for regression that leads to 
trustworthy and more accurate model predictions.

\begin{figure}[tb]                                                          
\hspace{26pt}    
\begin{minipage}[b]{.1\linewidth}         
  \centering      
  \centerline{\epsfig{figure=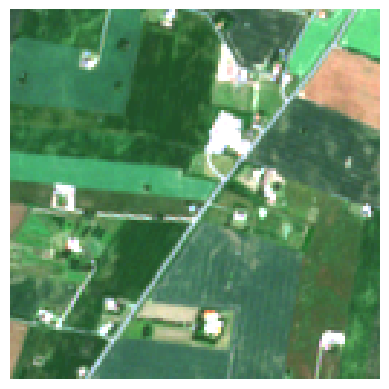,height=2.945cm,width=2.945cm}}           
  \vspace{0.001cm}   
  \centerline{\small a) Input image~~}\medskip  
\end{minipage}
%\hfill
\hspace{49pt}
\begin{minipage}[b]{0.1\linewidth}
  \centering 
  \centerline{\epsfig{figure=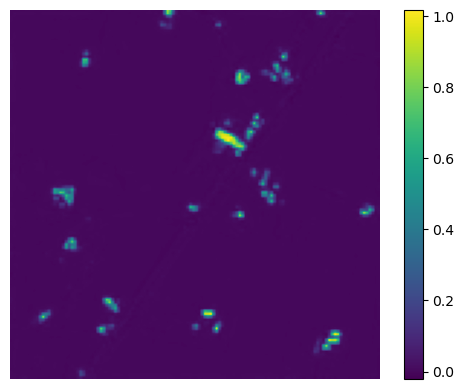,height=2.945cm,width=2.945cm}}      
  \vspace{0.001cm} 
  \centerline{\small b) Prediction~~}\medskip  
\end{minipage}
%\hfill 
\hspace{49pt}    
\begin{minipage}[b]{.1\linewidth}    
  \centering      
  \centerline{\epsfig{figure=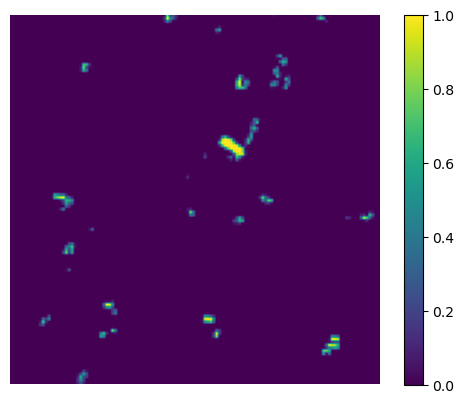,height=2.945cm,width=2.945cm}}      
  \vspace{0.001cm}    
  \centerline{\small c) Ground truth~~}\medskip 
\end{minipage}              

\hspace{26pt}   
\begin{minipage}[b]{.1\linewidth}           
  \centering         
  \centerline{\epsfig{figure=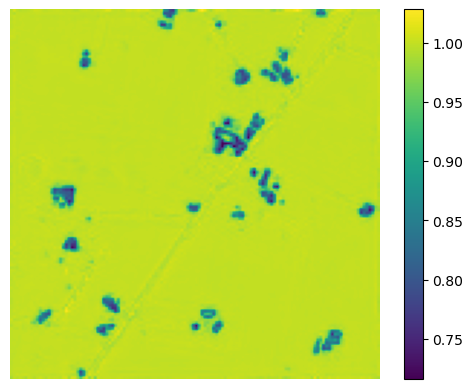,height=2.945cm,width=2.945cm}}    
  \vspace{0.001cm}  
  %\centerline{\small d) Incorrect~~}
  %\centerline{\small \, classifications~~}\medskip  
  \centerline{\small d) Confidence~~~}      
  \centerline{\small \, map, CARE (Ours)~~~}\medskip     
\end{minipage}
%\hfill 
\hspace{49pt}
\begin{minipage}[b]{0.1\linewidth}   
  \centering         
  \centerline{\epsfig{figure=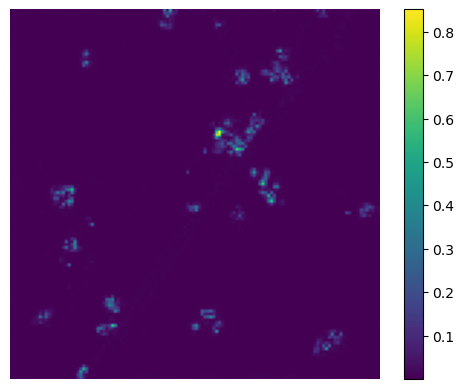,height=2.945cm,width=2.945cm}}       
  \vspace{0.001cm}    
  \centerline{\small e) Abs. error btw~~~}      
  \centerline{\small prediction \& gt~~~}\medskip   
\end{minipage} 
%\hfill  
\hspace{49pt}    
\begin{minipage}[b]{.1\linewidth}                
  \centering                        
  \centerline{\epsfig{figure=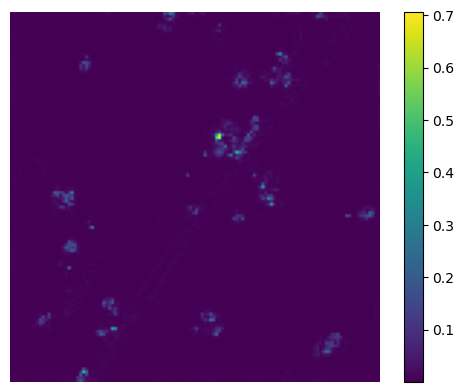,height=2.945cm,width=2.945cm}}                    
  \vspace{0.001cm}     
  \centerline{\small f) Abs. error btw~}           
  \centerline{\small pred. uncertainty \& (e)~}\medskip        
\end{minipage}    
\vspace{-9pt}        
%\caption{\small Semantic segmentation and confidence assignment and assessments by CAS on Sentinel-2 \textit{multi-spectral} data using WorldCover.}                                                              
%\caption{\small Semantic segmentation and confidence assignment and assessments by CAS on Sentinel-2 \textit{multi-spectral} data using WorldCover.}                                                    
\caption{\small Regression and \textit{confidence} quantification: CARE using \eqref{eq:newequ1}-\eqref{eq:confidequation}.}                                                         
% Pixel-wise regression and confidence estimation and assessment by the model CARE described in Sec.~\ref{sec:secmethod} on Sentinel-2 data.  
%Semantic segmentation, confidence estimation, and confidence assessments by our model CAS described in Sec.~\ref{sec:sectionProposedMethod} in the subsection ``Model CAS'' on Sentinel-2 images using the dataset \textit{WorldCover}      
\label{fig:figure4CAS3}                             
\label{fig:figure3CAS3}      
\end{figure}

\subsection{Baseline model n-shot evaluation, Table III}   
\noindent In Table~\ref{tab:table2resultsCASmain22333}, we evaluate the model Error-Sorting Confidence \cite{Confquantification1} on the Sentinel-2 PhilEO Bench building density estimation downstream task \cite{Fibaek1}, to \textit{compare} these results with the results of our proposed model CARE in Table~\ref{tab:table2resultsCASmainn22222}.       
Error-Sorting Confidence is used in Table~\ref{tab:table2resultsCASmain22333} in the \textit{fine-tuning} stage, i.e. when starting from and using a trained EO FM \cite{Fibaek1, EGU1}.        
Here, Error-Sorting is also an ablation study, i.e. using only the second term in \eqref{eq:confidequation}. 
We observe that for $n=100$ training samples per region, the percentage improvement of CARE compared to Error-Sorting Confidence is $8.93\%$ for the MSE, and $73.47\%$ for the median error.

\subsection{Qualitative evaluation of CARE}      
\noindent For the evaluation of 
%the model        
CARE, we also examine qualitative results, in Figs.~\ref{fig:figure4CAS}-\ref{fig:figure3CAS3}.                       
%In particular, w
%We perform qualitative evaluation and verify that the confidence estimate is correct.            
%That is, the assigned confidence has to be meaningful and has to also be consistent with reality, as well as with the correctness and accuracy of the model regression output.   
In addition to the numerical evaluation metrics presented in the preceding subsections, we also evaluate the proposed model CARE qualitatively.  
For the error in the prediction, i.e. in (e), the \textit{variation}      in the regression output is, for example, because we underestimate in (b) the building density.   
In (f), the predicted uncertainty is $(1-\text{confidence})$, i.e. using (d).    
In the colorbar in (f), 
%on the right, 
the maximum value is \textit{smaller} than $1$, and this is desirable, i.e. we have \textit{less}      highlighted and emphasized yellow color.

\section{Conclusion}  
\noindent In this paper, we have proposed the model CARE for pixel-wise regression tasks, for building density estimation from satellite Sentinel-2 data. 
Our model CARE computes and assigns a regression measure and a confidence metric to the output results of deep neural networks that solve regression problems. 
CARE not only predicts building density effectively but also provides reliable confidence scores that indicate trust in its predictions. 
The results reveal that CARE achieves a percentage improvement over Gaussian-Output NLL and other confidence estimation models. 
Evaluation metrics include the MSE, the average absolute error and the median absolute error, allowing comprehensive performance analysis against several baseline models. 
The experiments demonstrate that CARE significantly outperforms other baseline methods, showing improvements in various metrics. 
CARE shows a notable reduction in prediction error and demonstrates a strong correlation between confidence values and prediction errors.
The results summarize the innovation introduced by the proposed model CARE, underscoring its dual capability of providing both accurate regression outputs and meaningful confidence metrics. 
We have shown that confidence quantification and assessment is critical for the deployment of models in real-life applications, particularly in EO tasks where uncertainties can have significant implications.
By addressing the gap in confidence estimation for regression tasks in EO Foundation Models for building density estimation, this work contributes a valuable methodology and insights that can further enhance the practical application of neural networks in geospatial analysis. 
This underscores the importance of developing models that not only provide predictions but also quantify certainty, facilitating informed decision-making in critical areas such as urban planning and environmental monitoring.
%The work stands to benefit both researchers developing advanced algorithms and practitioners seeking robust models for interpretation and application of EO data.

%using the average absolute error between the absolute error of the regression output and the metric $(1-\text{confidence})$.   
%We have demonstrated that CARE model is effective and is able to both predict building density and estimate confidence.    
%In addition, we have also compared our model with other baseline models and we have shown that the proposed model CARE outperforms other approaches.

%to the best of the authors' knowledge
%first time
%confidence
%regression
%applied in earth observation and remote sensing, and specifically for building density estimation from satellite Sentinel-2 multi-spectral data

%\section{REFERENCES}

%Parameter-Efficient Fine-Tuning for Foundation Models, 2025

%Directional Gradient Projection for Robust Fine-Tuning of Foundation Models, 2025

%Confidence Estimation via Sequential Likelihood Mixing, 2025

\end{document}